\documentclass[11pt]{article}

\usepackage[utf8]{inputenc}
\usepackage[T1]{fontenc}
\usepackage[a4paper,margin=1in]{geometry}
\usepackage{times}
\usepackage{microtype}
\usepackage{setspace}
\usepackage{natbib}
\usepackage{hyperref}
\hypersetup{
    colorlinks=true,
    linkcolor=blue,
    citecolor=blue,
    urlcolor=blue
}

\usepackage{mathtools}
\usepackage{multirow}
\usepackage{rotating}
\usepackage{caption}
\usepackage{subcaption}

\usepackage{algorithm2e}
\usepackage{xcolor}
\usepackage{todonotes}

\title{\LARGE \bf
Overcoming the Generalization Limits of SLM Finetuning for Shape-Based Extraction of Datatype and Object Properties
}

\author{
Célian Ringwald, Fabien Gandon, Catherine Faron, Franck Michel, Hanna Abi Akl\\[4pt]
\small Univ. Côte d’Azur, Inria, CNRS, I3S, Sophia-Antipolis, France\\
\small \texttt{name.lastname@inria.fr}
}

\date{October 15, 2025}

\begin{document}
\maketitle

\begin{abstract}
Small language models (SLMs) have shown promises for relation extraction (RE) when extracting RDF triples guided by SHACL shapes focused on common datatype properties. This paper investigates how SLMs handle both datatype and object properties for a complete RDF graph extraction. We show that the key bottleneck is related to long-tail distribution of rare properties.
To solve this issue, we evaluate several strategies: stratified sampling, weighted loss, dataset scaling, and template-based synthetic data augmentation.
We show that the best strategy to perform equally well over unbalanced target properties is to build a training set where the number of occurrences of each property exceeds a given threshold.
To enable reproducibility, we publicly released our datasets, experimental results and code. Our findings offer practical guidance for training shape-aware SLMs and highlight promising directions for future work in semantic RE.
\end{abstract}


\section{Introduction}

Relation Extraction (RE) is a core task in information extraction, which involves identifying entities in unstructured text and classifying semantic relationships between them. It plays a crucial role in populating and enriching knowledge graphs (KGs) by transforming textual data into structured relational triples. Recent advances in NLP, notably with language models (LMs), have significantly improved RE systems. These models now offer more context-aware and adaptable extraction capabilities with minimal task-specific supervision. 
To date, the primary focus of RE research has been on frequently occurring relations and well-defined entity types. Canonical datasets~\cite{zhang2017tacred,elsahar-etal-2018-rex,yao-etal-2019-docred} were introduced to support this direction at scale, often employing distant supervision to align textual content with existing KGs.
Such approaches have yielded proficient models in extracting various relational facts encompassing heterogeneous entity classes and relation types within a unified framework. Nonetheless, extracting less frequent and more complex relational patterns remains challenging for LMs. 

\noindent \textbf{Task definition:} Following the formulation established in~\cite{DBLP:conf/esws/RingwaldGFMA25}, we propose learning RDF pattern extractors $\mathcal{M}_{Db}$ by finetuning Small LMs (SLMs). Training such models relies on a dual base $Db\subseteq$ $W\times{G}$, where $W$ is a set of texts and $G$ the set of associated KGs. \\
The target of the extraction is given by a SHACL shape $s$. The extractor is defined as a function $E_{Db}$: $W \times S \rightarrow G; (t,s) \mapsto \hat{g}$, 
where $t \in W$ is an input text, $s \in S$ is a set of SHACL shapes, and $\hat{g}$ is an RDF graph implied by $t$ and valid against $s$. We derive from $s$ the set $\Pi$ of all possible patterns (i.e. combinations of properties from that shape).
We call \textit{example-specific pattern} each such property combination $\pi$ in $\Pi$. An RDF graph $g$ is valid against a pattern $\pi$ if it contains exactly all the properties in $\pi$ and no more.

In~\cite{DBLP:conf/esws/RingwaldGFMA25} we focused on the extraction of datatype properties. In this paper, we extend the pattern-based RDF extraction to SHACL shapes that also contain object properties. 
Moreover, we highlight the critical challenge of ensuring balanced performance across all properties defined within a given SHACL shape, including under-represented properties, and we propose several solutions to address this issue.  
To summarize, this paper addresses the two following research questions: 
\begin{itemize}
    \item \textbf{RQ1.} Is it possible to perform pattern-based RE for both datatype and object properties?
    \item \textbf{RQ2.} How do finetuned shape-based relation extractors perform on datasets with unbalanced property 
    distributions?
\end{itemize}

\section{Related Work}\label{sec:related_work}

\subsection{Ontology-Driven Extraction} 
Several approaches proposed guiding RE with a predefined schema or ontology. They enable the extraction of relations not merely as independent triples, but within a contextual framework considering typed entities and the domain and range of the targeted relations. 
An example would be specifically targeting DBpedia relations involving entities belonging to the DBpedia class \texttt{dbo:Company}.
An illustration of this paradigm is the SPIRES framework~\cite{10.1093/bioinformatics/btae104} which supports the construction of prompts based on schema-aware information retrieval. 
Among the various corpora developed for the RE task, Text2KGBench~\cite{10.1007/978-3-031-47243-5_14}  stands out as one of the few datasets explicitly designed to support ontology-driven RE. 

\textbf{Positioning:} Our previous work on RE from Wikipedia~\cite{DBLP:conf/esws/RingwaldGFMA25} aligns with these works, using SHACL shapes to represent ontology constraints. This showed encouraging results when finetuning SLMs to create RDF pattern-based relation extractors. However, it was limited to extracting datatype properties. In this paper we extend our pattern-based RE approach to the extraction of both datatype and object properties.

\subsection{Dealing with Under-represented Knowledge} 
Seq2Seq models finetuned on massive datasets, such as REBEL~\cite{huguet-cabot-navigli-2021-rebel-relation} and GenIE~\cite{josifoski-etal-2022-genie}, generally offer good performance but are not well-suited to deal with rare properties, as shown in~\cite{he-etal-2025-language,josifoski-etal-2023-exploiting}. The authors of SynthIE attempt to address this issue by generating synthetic data using a Large Language Model (LLM) as a data generator, which is subsequently used to train an SLM. Despite the good performance of these models on synthetic test data, their results do not reach the same level of performance on real-world, noisy data, such as those covered by REBEL. This limitation becomes obvious when evaluating SynthIE's performance on the long-tail distribution of properties in the REBEL dataset. 

Text2KGBench~\cite{10.1007/978-3-031-47243-5_14} was initialized using two LLMs (Vicuna and Alapaca), and the results show the difficulty of extracting graphs described by an ontology.
The main explanation is the models' inability to handle numerous under-represented properties. In such cases, using the macro F1 metric is crucial to highlight the performance disparities across properties; unfortunately, this analysis is rarely conducted in RE. 
Similarly, \cite{arzt2025relationextractionpatternmatching} compares the performances of LMs on a relation classification task focusing on biographic data. It shows LM limitations on extracting under-represented relations, and highlights the importance of cross-dataset evaluation. 
Considering biographic data, \cite{AL2024-knowledge3} emphasises the importance of scaling laws in learning a specific \textit{piece of knowledge} and shows that minimal exposure to a rare case is required for good performance, i.e., a minimum number of times a given piece of knowledge is represented in the training set. More precisely, the experiments show that 100 exposures during a training process are insufficient, and 1,000 exposures is proposed as a minimal condition to learn a fact. However, these conclusions were drawn in the context of pretraining models for memorization. In our context, we are finetuning models to retrieve facts from text. 
Among the numerous insights from this work, we also note the finding that data augmentation does not harm—and may even improve—model performance. This observation supports the potential of synthetic data augmentation strategies~\cite{ding2024dataaugmentationusinglarge}, especially in low-resource scenarios. 

To address data imbalance, a standard solution in machine learning is using re-weighted loss functions to inversely balance each class by its importance in the training set~\cite{8237586,Jiang_Chan_Xue_Liu_Guo_2025,xu-etal-2022-towards-realistic}. However, only little work investigates it in the context of LMs and RE.

\textbf{Positioning:} Before we can generalize RDF pattern-based RE approaches to a broad set of SHACL shapes, we must first address the challenge of handling the long-tail distribution of properties 
with a single shape. To this end, we investigate stratification and re-weighted loss techniques, as well as several training data construction strategies—including scaling, augmentation, and ensuring sufficient exposure per property. The latter allows us to verify whether the ``sufficient exposure'' hypothesis proposed by~\cite{AL2024-knowledge3} at the fact level also applies at the property level.

\subsection{Contributions} 
(1) We extend the RDF pattern-based relation extraction in~\cite{DBLP:conf/esws/RingwaldGFMA25} to both datatype and object properties. 

\noindent
(2) We propose and benchmark a wide number of solutions, revealing the need to respect a minimal exposure threshold to deal with the long-tail distribution of properties. 

\noindent
(3) The produced code\footnote{\href{https://github.com/datalogism/Kastor}{Kastor github}} and datasets\footnote{\href{https://zenodo.org/records/15917325}{Link to Zenodo repository}} are open and reusable.

\section{Extracting Object and Datatype Properties}\label{sec:part1}

\subsection{Knowledge Distillation}
In~\cite{DBLP:conf/esws/RingwaldGFMA25}, we considered the distillation of an initial \textbf{dual base $\mathcal{K}$} consisting of the 2022.09 DBpedia datadump\footnote{\href{https://databus.dbpedia.org/cringwald/collections/kstor}{DBpedia data dump used}} which contains 6,109,994 Wikipedia abstracts and their DBpedia graphs:
\begin{equation} \label{eq:K}
\begin{split}
\mathcal{K} := \{(w,g) \in \mathcal{W}\times\mathcal{G}, \exists \, e \in IRI \text{ such that } \\ desc_{\mathcal{W}}(e)=w \: \wedge \: desc_{\mathcal{G}}(e)=g\}
\end{split}
\end{equation}

Here we extend the maximal shape defined in ~\cite{DBLP:conf/esws/RingwaldGFMA25}\footnote{\href{https://github.com/datalogism/Kastor/blob/main/shapes/PersonShape_op_and_dp.ttl}{dbo:Person shape containing datatype and object prop.}} that considered only the following set of datatype properties:\\
$\mathcal{P}(s_{dt}^*)=$ \begin{small}\{$\texttt{rdfs:label}, \texttt{dbo:alias}, \texttt{dbo:birthName},\texttt{dbo:birthDate},\\\texttt{dbo:birthYear},\texttt{dbo:deathDate},\texttt{dbo:deathYear}$\}\end{small}.\\
We now additionally consider the three following object properties commonly used to describe \texttt{dbo:Person} instances: \\
$\mathcal{P}(s_{op}^*)=$ \begin{small}\{$\texttt{dbo:birthPlace}, \texttt{dbo:nationality}, \texttt{dbo:deathPlace}$\}\end{small}\\
Our new maximal shape includes both types of properties:
$$\mathcal{P}(s^*)=\mathcal{P}(s_{dt}^*) \cup \mathcal{P}(s_{op}^*)$$

As done in~\cite{DBLP:conf/esws/RingwaldGFMA25}, we perform a data enrichment aiming to compensate for potentially missing datatype property values with the following set of rules\footnote{For details \href{https://github.com/datalogism/Kastor/blob/main/kstor/rules/Person_datatypes_rules_NS.rul}{Rules related to datatype prop.}}:
\begin{equation}
\mathcal{R}_{dt}: \begin{dcases}
       \texttt{dbo:deathDate}  \models \texttt{dbo:deathYear} \\
       \texttt{dbo:birthDate}  \models \texttt{dbo:birthYear}    
\end{dcases}
\end{equation}
Similarly, to improve the coverage of object properties, we introduce a new set of rules $\mathcal{R}_{op}$ enabling to infer countries from the objects of properties \texttt{dbo:birthPlace}, \texttt{dbo:deathPlace} and \texttt{dbo:nationality}\footnote{For details \href{https://github.com/datalogism/Kastor/blob/main/kstor/rules/Person_dataobj_rules_NS.rul}{Rules related to object prop.}}:
\begin{equation}
\mathcal{R}_{op}: \begin{dcases}
      (s,p,o) \wedge (o,\texttt{dbo:country},c)  \models  (s,p,c) 
\end{dcases}
\end{equation}
Applying the rules $\mathcal{R}=\mathcal{R}_{dt} \cup \mathcal{R}_{op}$ to our initial knowledge graph produces an augmented version thereof:

\begin{equation} \label{eq:K-R-entails}
\begin{split}
\mathcal{K}^{\mathcal{R}} := \{(w,g')\, , \; (w,g) \in \mathcal{K}\, \\ \text{ where } g' \text{ is the result of applying } \mathcal{R} \text{ to } g \}
\end{split}
\end{equation}

The \textit{Wikicheck} post-processing step of the knowledge distillation retains only the triples that are supported by evidence from the corresponding Wikipedia abstract. In our earlier work, this validation step was limited to datatype properties whereby we verified the presence of the described values in the Wikipedia abstracts ($\mathcal{W}_{plain}$), using plain text search for string data and specific date parsers for dates.
However, "Wikichecking" the object properties is more challenging, as their values must be identified by DBpedia URIs which are not included in the plain text abstracts. 
To support this, we retrieve Wikipedia pages' HTML representation we use the Wikipedia's REST API\footnote{\url{https://en.wikipedia.org/api/rest_v1/}}, and ensure temporal consistency of the pages with the graph using DBpedia by matching each page to its corresponding \texttt{dbo:wikiPageID}, and convert the obtained HTML abstracts to Markdown ($\mathcal{W}_{MD}$).
Using $\mathcal{W}_{MD}$, we then extend Wikicheck to object properties by verifying whether the DBpedia URIs proposed as object values are present in the Markdown links. Additionally, to reduce potential noise, we verify for each object property whether the type of the object—retrieved from DBpedia using its URI—is valid with respect to the property's expected range as defined in the SHACL shape. Due to Wikimedia's API rate limits, we apply this process only to a random sub-sample of dbo:Person entities to avoid selection bias. 
As a result of the knowledge distillation process of the dual base describing  \texttt{dbo:Person} instances with the shape $s^*$, we obtained a distilled version of the base $\mathcal{K}_{dist}(s^*) $ containing 136,718 graphs:\\

\begin{equation} \label{eq:K-s-star_RW_obj}
\begin{split}
\mathcal{K}_{dist}(s^*)  \coloneqq \{ (w,w_{MD},g) ; (w,g) \in \mathcal{K}^{\mathcal{R}}  \, , \; w_{MD} \models g, \\
\exists \pi \in \Pi(s^*) \ ; g \text{ is valid against } \pi \}
\end{split}
\end{equation}

where $w_{MD}$ is the markdown version of $w$.

\subsection{Datasets}\label{sec:p1_dataset} 

$\mathcal{K}_{dist}(s^*)$ is used to sample training sets to finetune pre-trained langague models. Since our finetuning rely on the BART model, that was released in 2021,  we paid close attention to
select only Wikipedia articles
published strictly before this date. We also intentionally biased the training set to ensure that each graph would contain at least one datatype property and one object property. We denote this dataset $D1$:
\begin{equation} \label{eq:dataset_biased_old}
\begin{split}
D1 := \{ (w,w_{MD},g) \in \mathcal{K}_{dist}(s^*) ; \\
g \text{ contains at least one property of } s_{op}^*\\
\text{and at least one property of } s_{dt}^*\\
\text{and } w \text{ was created before 2021} \}
\end{split}
\end{equation}

Table~\ref{tab:dataset_stat} provides the number of triples for each property, as well as its frequency in the description of instances of \texttt{dbo:Person}, in the distilled base $\mathcal{K}_{dist}(s^*)$ and in the sample $D1$. 
\begin{table}[h!]
\caption{Frequency and number of triples per property in the distilled base and in sample $D1$}
\label{tab:dataset_stat}
\centering
\resizebox{0.7\linewidth}{!}{
\begin{tabular}{l|rr|rr} & \multicolumn{2}{c|}{$\mathcal{K}_{dist}(s^*)$} & \multicolumn{2}{c}{$D1$} \\ \hline
 prop        & No.                        & Freq                  & No.      & Freq  \\ \hline
 birthYear   & 105,818                   & 0.77                  & 1,010   & 0.84           \\
                          birthPlace  & 15,279                    & 0.11                  & 996     & 0.83           \\
                           birthDate   & 97,039                    & 0.71                  & 927     & 0.77           \\
                           label       & 92,691                    & 0.68                  & 870     & 0.73           \\
                           deathYear   & 37,633                    & 0.28                  & 457     & 0.38           \\
                           deathDate   & 32,743                    & 0.24                  & 395     & 0.33           \\
 deathPlace  & 4,859                     & 0.04                  & 291     & 0.24           \\
                           nationality & 2,211                     & 0.02                  & 162     & 0.14           \\
                           birthName   & 12,265                    & 0.09                  & 130     & 0.11           \\
                           alias       & 1,398                     & 0.01                  & 14      & 0.01           \\ 
\end{tabular}
}
\end{table}

\subsection{REBEL as Baseline}
To compare our method to the state of the art, we chose to compare our results with those of REBEL-large, a model with 406 M parameters, both a strong baseline today and a major inspiration for our approach.
As REBEL was not specifically designed for the extraction of RDF triples or structured text, we made several adaptations to ensure a fair comparison.
First, we aligned the properties of Table~\ref{tab:mapping_rel1} used by REBEL with the corresponding DBpedia properties. The resulting shape is denoted $s^*_{rebel}$.
Then we linked the subject and object of the object properties retrieved by REBEL to DBpedia resources using DBpediaLookup~\footnote{\url{https://lookup.dbpedia.org/}}. Finally, 
we enriched REBEL's output with the triples that could be inferred by the rules $\mathcal{R}_{dt}$ and $\mathcal{R}_{op}$.

\begin{table}[h!]
\centering
\caption{REBEL/DBpedia relation alignment}\label{tab:mapping_rel1}
\resizebox{0.7\linewidth}{!}{
\begin{tabular}{l|l}
REBEL & DBpedia \\
\hline
    "place of birth" &  \texttt{dbo:birthPlace} \\
    "place of death" &  \texttt{dbo:deathPlace} \\
    "country of citizenship" & \texttt{dbo:nationality} \\
    "date of birth" & \texttt{dbo:birthDate} \\
    "date of death" & \texttt{dbo:deathDate} \\
\end{tabular}
}
\end{table}

\subsection{Training Details}\label{sec:p1_details_ft} 
All the trained models follow the configuration described in our previous work~\cite {DBLP:conf/esws/RingwaldGFMA25}. We used the BART-base (140M parameters) pre-trained model. We linearised the graphs following the ``TurtleLight one line and factorised''~\cite {DBLP:conf/esws/RingwaldGFMA25} syntax which demonstrated the best performance. The models were finetuned on a single Nvidia Tesla V100 GPU using the same configuration: an inverse square root scheduler with an initial learning rate of 0.00005, 1,000 steps of warmup, and configured with an early stop mode with patience of 5 steps. 
We used the same prompt to finetune the models: ``\texttt{\$entity\_URI : \$Abstract}'' where \$Abstract is a Wikipedia abstract and \$entity\_URI the URI of the corresponding entity in DBpedia. Each model was finetuned using a 10-fold cross-validation.

We trained models for each combination of Wikipedia abstract type $W_{type}\in \{MD, PLAIN\}$ (i.e. respectively using $w_{MD}$ or $w$ in $(w,w_{MD},g) \in \mathcal{K}_{dist}(s^*)$) 
and shape $s \in \{s^*,s_{op}^*,s_{dt}^*\}$:

\begin{equation} \label{eq:M'}
\begin{split}
\mathcal{M}D1(W_{type},s):\\
\begin{dcases}
\mathcal{K}_{dist}(s)\rightarrow \mathcal{G} \\
w \mapsto \hat{g} ; \hat{g}  \leftrightarrow \mathcal{P}(g) \wedge (w,w_{MD},g) \in D1 ; \text{if } W_{type}= PLAIN \\
w_{MD} \mapsto \hat{g} ; \hat{g}  \leftrightarrow \mathcal{P}(g) \wedge (w,w_{MD},g) \in D1 ; \text{if } W_{type}= MD \\
\end{dcases} 
\end{split}
\end{equation}
We also trained a model $\mathcal{M}D1(W_{PLAIN},s^*_{rebel})$ corresponding to the REBEL baseline, to compare with the above models.

\subsection{Metrics and Evaluation} 
To compare the performance of the obtained models, we consider using the strict comparison of the expected values with the generated ones to compute the macro-F1 ($F_1^+$) and the micro-F1 ($F_1^-$). $F_1^-$ is the average of the $F_1$ scores computed over all the graphs without considering the property distribution, while $F_1^+$ is the average of the $F_1$ scores computed per property.
The micro-F1 captures the overall performance but is biased towards more frequent properties, while the macro-F1 provides a more balanced view, especially in the presence of rare properties. Additionally, we compute the micro- recall and precision scores.

All trained models are evaluated across the 10 folds using the same test sets 
from $D1$, making the results comparable.
For each model, we report the mean and standard deviation of the metrics over the folds.  

\subsection{Results}

\begin{table}[h!]
\centering
\caption{Performance of $\mathcal{M}D1$ models finetuned with different shapes and input types}
\label{tab:xp2_all_old_res}
\resizebox{0.8\linewidth}{!}{
\begin{tabular}{l|ll|ll}
 Model   & $Recall$   & $Prec.$ & $F_1^-$ & $F_1^+$ \\
\hline 
 $\mathcal{M}D1(MD,s^*_{dt})$ & 98 $\pm$ 2 & 89 $\pm$ 3 & 97 $\pm$ 3 &  89 $\pm$ 11 \\
 $\mathcal{M}D1(MD,s^*_{op})$ & 91 $\pm$ 13 & 88 $\pm$ 13 & 90 $\pm$ 13 & 83 $\pm$ 22 \\
 $\mathcal{M}D1(MD,s^*)$ & 95 $\pm$ 5 & 92 $\pm$ 6 & 94 $\pm$ 5 & 86 $\pm$ 14   \\
\hline
 $\mathcal{M}D1(PLAIN,s^*_{dt})$  & 98 $\pm$ 2 & 96 $\pm$ 3 & 98 $\pm$ 3 &  91 $\pm$ 10  \\
 $\mathcal{M}D1(PLAIN,s^*_{op})$ & 90 $\pm$ 14 & 87 $\pm$ 15 & 89 $\pm$ 14 & 83 $\pm$ 22  \\
 $\mathcal{M}D1(PLAIN,s^*)$ & 96 $\pm$ 4 & 94 $\pm$ 6 & 95 $\pm$ 5 & 88 $\pm$ 14 \\
\end{tabular}
}
\end{table}
The training details for the models referred to in this section are available on Wandb\footnote{\url{https://wandb.ai/celian-ringwald/GenLimits_Part1}}.
Table~\ref{tab:xp2_all_old_res} provides a summary of the metrics recorded for the different models. 

\subsubsection{Datatype vs. Object Properties}\label{subsec:prop_freq_sec}

We first observe that extracting from Markdown content does not help the model to extract object properties. Possibly, the large number of URIs embedded in the Markdown annotations creates noise.
Second, we observe that we achieve lower F1 scores on object properties ($s^*_{op}$) than on datatype properties ($s^*_{dt}$). Moreover, the high standard deviation recorded on $s^*_{op}$ further suggests that extracting object properties is less stable and more error-prone.
An explanation for the deterioration in performance on object properties could be related to a discrepancy in the distribution of properties targeted by the shape.
To investigate this hypothesis, we extended our analysis by distinguishing between frequent properties and rare properties.
A \textit{frequent property} is a property whose frequency is greater than the average property frequency $\mu_p$. Conversely, a \textit{rare property} has a frequency below $\mu_p$. 
In the case of $D1$, $\mu_p=0.295$. 
Thence, we derived two subsets of $s^*$:
\begin{itemize}
    \item $s^*_+$ contains only frequent properties:\\ $\mathcal{P}(s_{+}^*)=$ \begin{small}\{$\texttt{rdfs:label}, \texttt{dbo:birthDate}, \texttt{dbo:birthPlace},\\
    \texttt{dbo:birthYear}, \texttt{dbo:deathDate}, \texttt{dbo:deathYear}\}$\end{small} 
    \item $s^*_-$ contains only rare properties:\\ $\mathcal{P}(s_{-}^*)=$  \begin{small}\{$\texttt{dbo:birthName},\texttt{dbo:nationality}, \\
    \texttt{dbo:deathPlace}, \texttt{dbo:alias}\}$\end{small}
   
\end{itemize}
We finetuned two models on $D1$, one with only the properties in $s^*_-$, and one with only those in $s^*_+$.
The micro and macro F1 scores in Table~\ref{tab:xp3_all_old_res} confirm that the frequency of a property in the training set directly impacts the performance of the extractor. 
The last line of the table, with $s^*_-$, also indicates that specialising a model on rare properties is not sufficient to perform well.
\begin{table}[h!]
\centering
\caption{Performance of $\mathcal{M}D1$ models finetuned on the new proposed subsets}
\resizebox{0.5\linewidth}{!}{
\begin{tabular}{l|l|l}
\label{tab:xp3_all_old_res}
$Model$ &  $F_1^-$ & $F_1^+$ \\
\hline 
$\mathcal{M}D1(PLAIN, s^*)$ & 95 $\pm$ 5 & 88  $\pm$ 15  \\
\hline
$\mathcal{M}D1(PLAIN, s^*_+)$ & \textbf{96 $\pm$ 4} &  \textbf{96 $\pm$ 5}   \\
$\mathcal{M}D1(PLAIN, s^*_-)$ & \underline{70} $\pm$ 21 & \underline{64 $\pm$ 26} \\
\end{tabular}
}
\vspace{-10pt}
\end{table}

\subsubsection{Comparison with the REBEL baseline}

We first note that our models successfully produce 100\% of the triples in well-formed RDF, with all generated subjects being the expected URIs; by contrast, only 70\% of the subjects produced by REBEL and entity-linked are equals to the expected URIs. 
Table~\ref{tab:rebel_comparison} shows that despite decent accuracy, REBEL fails to extract many facts (as evidenced by the recall), and performs significantly worse than our models with a substantial gap between the micro and macro $F_1$, indicating that REBEL is only capable of extracting a small number of properties.
These results confirm that, overall, our shape-based models outperform REBEL, especially in terms of structural validity and generalization.

\begin{table}[h!]
\centering
\caption{Comparison of $\mathcal{M}D1$ with REBEL}
\label{tab:rebel_comparison}
\resizebox{0.8\linewidth}{!}{
\begin{tabular}{l|ll|ll}
Model    & $Recall$   & $Prec.$ & $F_1^-$ & $F_1^+$ \\
\hline 
$\mathcal{M}D1(PLAIN,s^*_{rebel})$ & 96 $\pm$ 5 & 93 $\pm$ 5 &  94 $\pm$ 5 & 89 $\pm$ 10 \\
 $REBEL$ & 54 $\pm$ 1 & 83 $\pm$ 1 & 65$\pm$ 1  & 41 $\pm$ 1 \\
 $REBEL\times DBpediaLookup$ & 59 $\pm$ 0.3 & 83 $\pm$ 1 & 69 $\pm$ 0.3  & 44 $\pm$ 0.6 
\end{tabular}
}
\end{table}

\section{Long-Tail Property Distribution Challenges}\label{sec:part2}
Further exploratory experiments conducted on a larger scale on 16 shapes derived from the Text2KGBench micro-ontology\footnote{\url{https://wandb.ai/celian-ringwald/Text2KGBench_KastorModels}} also recorded the same high F1 deviations, which can also be explained by the unbalanced distribution of properties (see Section~\ref{sec:part1}). To address this issue, we propose and investigate in this section several solutions, while keeping the focus on the \texttt{dbo:Person} shape which is well described in our distilled Knowledge Graph: (1) we increase the size of the training sets and construct a sample over-representing rare properties; (2) we use stratified sampling and weighted loss function during the finetuning. 
Additionally, we construct four datasets with varying property distributions to cross-evaluate the resulting models and assess their ability to generalize.

\subsection{Baselines}

\subsubsection{Training at scale and oversampling rare properties}
Until now, we have focused on finetuning the model using small datasets, which previously offered a good trade-off between performance and resource consumption. However, a well-established solution in deep learning is to scale the training data. 
To evaluate the effect of the training data size on performance, we trained new models on three new datasets:  $D2$, which contains 2,400 examples (twice $D1$), $D4$, containing 4,800 examples, and finally $D10$, containing 12,000 examples. In addition, we finetuned a model on another type of sample, $D_-$, which contains for each example at least one of the properties of $s^*_-$. The distribution of the properties of all these datasets is detailed in Table~\ref{tab:dataset_stat_2}.

\subsubsection{Working with a stratified set}\label{sec:part_strat}
The imbalanced nature of the data used for training our models in a 10-fold cross-validation process can affect the performances recorded in two key ways: (1) a training fold may not contain any example of a rare property, which will dramatically affect the performance of the model being trained; and (2) the validation and test folds may also lack rare properties, potentially leading to an overestimation or underestimation of the model’s true performance.
Our RE task aims to extract multiple properties related to a unique entity, which entails an overlapping stratification since a given graph can be classified with respect to multiple properties. To break this overlap, a solution is to consider each combination of properties as a class. But the number of patterns that can be derived from our shape is too large to allow direct stratification, even if we consider only the $|\Pi_{\mathcal{K}_{dist}}|=331$ distinct patterns realized in our distilled knowledge graph.
Therefore, we propose to focus the stratification strategy on the subset of rare properties defined in~\ref{subsec:prop_freq_sec}.

This strategy, further detailed on our github\footnote{\url{https://github.com/datalogism/Kastor/blob/main/doc/SamplingStrategies.md}}, is implemented by three rules: (1) if a graph contains exactly one rare property, it is assigned to the stratum corresponding to that property; (2) if it contains several rare properties, it is assigned to the stratum of the least represented rare property so far in the sample being constructed; (3) if a graph contains no rare property, it is assigned to the “Other” stratum. In order to use the same overall number of examples per model and increase our chances of obtaining rare properties in each fold, we train these models using 5-folds.

\subsubsection{Property-based weighted loss}
Until now, the finetuning of our models was based on minimizing the Cross-Entropy (CE) Loss between the linearized expected graph (${y}$) and the predicted one ($\hat{y}$). 
To adapt to the seq-to-seq context, the CE typically averages the token level cross-entropy over the maximum output length $T$.
\begin{equation} \label{eq:LOSSCE}
Loss_{CE}({y},\hat{y})=-\frac{1}{|T|}*\sum_{t=1}^{|T|}{y}_t\log(\hat{{y}_t})
\end{equation}

A common approach to address class-imbalanced data in machine learning is to use a weighted loss assigning different weights to the training instances according to their class, in order to better consider under-represented ones. In our context, it means being able to balance each linearized graph with a chosen weight $\omega_j$.

\begin{equation} \label{eq:LOSSWCE}
Loss_{WCE}({y},\hat{y})= \omega_{y}.Loss_{CE}({y},\hat{y})
\end{equation}
We reuse the stratification defined in Section~\ref{sec:part_strat} as a classification system to derive the weights from the cardinality of each stratum. Since each graph $y$ is associated to a stratum $c_y \in C$, we use the cardinality of each stratum $c_i$ noted here $|c_i|$, to compute the weight $w_y$ associated to a given sequence $y$. It is an inverse-log frequency that gives more importance to examples associated with less represented strata:
\begin{equation} \label{eq:omega}
\omega_y=log(\frac{\sum^{|C|}_{i=1}|c_i|}{|c_y|})
\end{equation}

\subsection{Cross-Evaluation}
Instead of a 10-fold cross-validation approach, we propose to test the configurations described in the previous section using four independent datasets and containing none of the training data. Each of them contains 200 examples and is designed to assess specific generalization aspects in the same conditions for all the models: 
\begin{itemize}
    \item $d_N$ (Never seen examples): BART was pre-trained on a full Wikipedia dump. This dataset gathers articles published after BART’s release. It allows us to evaluate how the model performs on content not seen during either pre-training or finetuning.
     \item $d_+$ (Frequent properties): This dataset consists of randomly sampled $\mathcal{K}_{dist}(s^*)$ examples including only frequent properties ($s^*_+$).
    \item $d_-$ (Rare properties): This dataset is composed of $\mathcal{K}_{dist}(s^*)$ examples including at least one rare properties ($s^*_-$).
    \item $d_R$ (Random properties): This dataset contains $\mathcal{K}_{dist}(s^*)$ random examples regardless of the article’s publication date and the frequency of properties used.
\end{itemize}

\subsection{Results}
\begin{table*}\centering
\caption{F1 scores obtained by each model on the four cross-evaluation datasets. Metrics are averaged over all folds and the standard deviation is always inferior to 5pt. Best scores are in bold and worst ones are underlined.}\label{tab:Res_all}

\resizebox{\linewidth}{!}{
\begin{tabular}{c|l|rrlr|llll}
\multirow{2}{*}{Model} & \multicolumn{1}{c|}{\multirow{2}{*}{Config}} & \multicolumn{4}{c|}{$F_1^-$}                                                               & \multicolumn{4}{c}{$F_1^+$}   \\ \cline{3-10} 
                       & \multicolumn{1}{c|}{}                        & \multicolumn{1}{l}{$d_N$} & \multicolumn{1}{l}{$d_+$} & $d_-$ & \multicolumn{1}{l|}{$d_R$} & $d_N$ & $d_+$ & $d_-$ & $d_R$ \\ \hline
$\mathcal{M}D1(PLAIN, s^*)$       & CE (10-folds)                                & 82.31                     & 91.71                     & \underline{76.90} & 83.85                      & 63.07 & 96.72 & \underline{59.19} & 61.98 \\
$\mathcal{M}D_-(PLAIN, s^*)$       & CE (10-folds)                                & \underline{72.66}                     & \underline{76.77}                     & \textbf{95.09} & \underline{75.49}                      & \underline{57.48} & \underline{84.86} & \textbf{90.68} & \underline{58.13} \\
$\mathcal{M}D2(PLAIN, s^*)$       & CE (10-folds)                                & 82.08                     & 92.00                     & 77.57 & 84.58                      & 62.90 & 96.65 & 61.10 & 62.66 \\
$\mathcal{M}D4(PLAIN, s^*)$       & CE (10-folds)                                & 84.19                     & 91.88                     & 79.20 & 86.39                      & 64.18 & \textbf{97.06} & 63.34 & 65.68 \\
$\mathcal{M}D10(PLAIN, s^*)$    & CE (10-folds)                                & \textbf{84.42}                     & \textbf{91.94}                     & 86.05 & \textbf{86.81}                      &  \textbf{67.05} & 97.05 & \textbf{77.26} & 66.52 \\ \hline
$\mathcal{M}D10_{Strat}(PLAIN, s^*)$    & CE-STRAT (5-folds)                           & 83.87                     & 90.06                     & 83.56 & 86.45                      & 65.77 & 95.29 & 71.92 & \textbf{67.83} \\
$\mathcal{M}D10_{StratWCE}(PLAIN, s^*)$    & WCE-STRAT (5-folds)                          & 84.16                     & 91.26                     & 84.53 & 86.49                      & 66.02 & 96.20 & 74.93 & 66.51
\end{tabular}
}
\end{table*}

The training details for the models referred to in this section are available on Wandb\footnote{Model training: \url{https://wandb.ai/celian-ringwald/GenLimitsPart2_train},\\ Cross-Evaluation:\url{https://wandb.ai/celian-ringwald/GenLimitsPart2_crossEval}}.
Table~\ref{tab:Res_all} reports the experimental results.
The first conclusion that we can drawn is that neither the stratification strategy nor the weighted loss significantly boost performance. 
Unsurprisingly, increasing the size of the training set improves performance. This leads to nearly a 10-point increase on both $F_1^-$ and $F_1^+$ on the dataset of rare properties  ($d_-$), while also improving results on all other test configurations.
Moreover, all the configurations perform well on the dataset of frequent properties ($d_+$), and
testing the models on previously unseen documents ($d_N$) does not significantly affect their performance.

\begin{figure}[ht]
    \centering
    \caption{Averaged micro F1 by property and model}
    \label{fig:cross-dataset_comp}
    \begin{subfigure}[b]{0.48\linewidth}
        \centering
        \includegraphics[width=\linewidth]{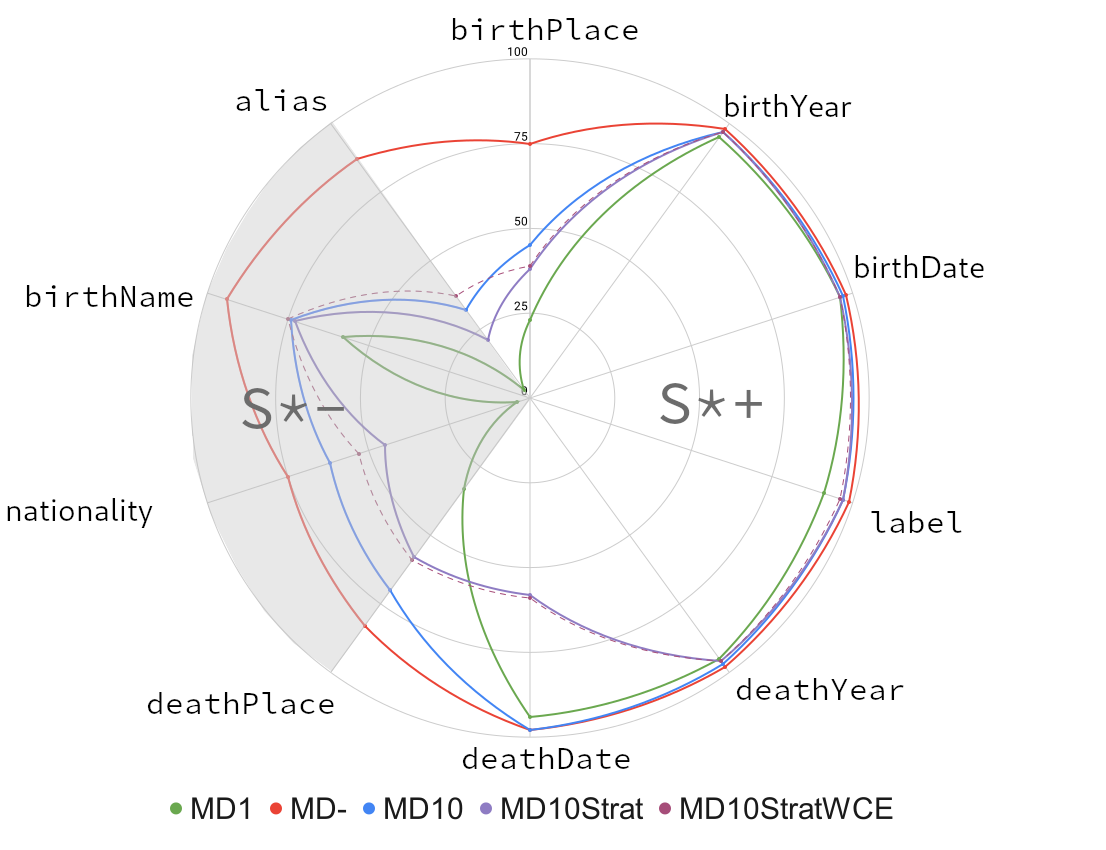}
        \caption{Evaluated on the rare property dataset $d_-$}
        \label{fig:D_-_step1_2}
    \end{subfigure}
    \hfill
    \begin{subfigure}[b]{0.48\linewidth}
        \centering
        \includegraphics[width=\linewidth]{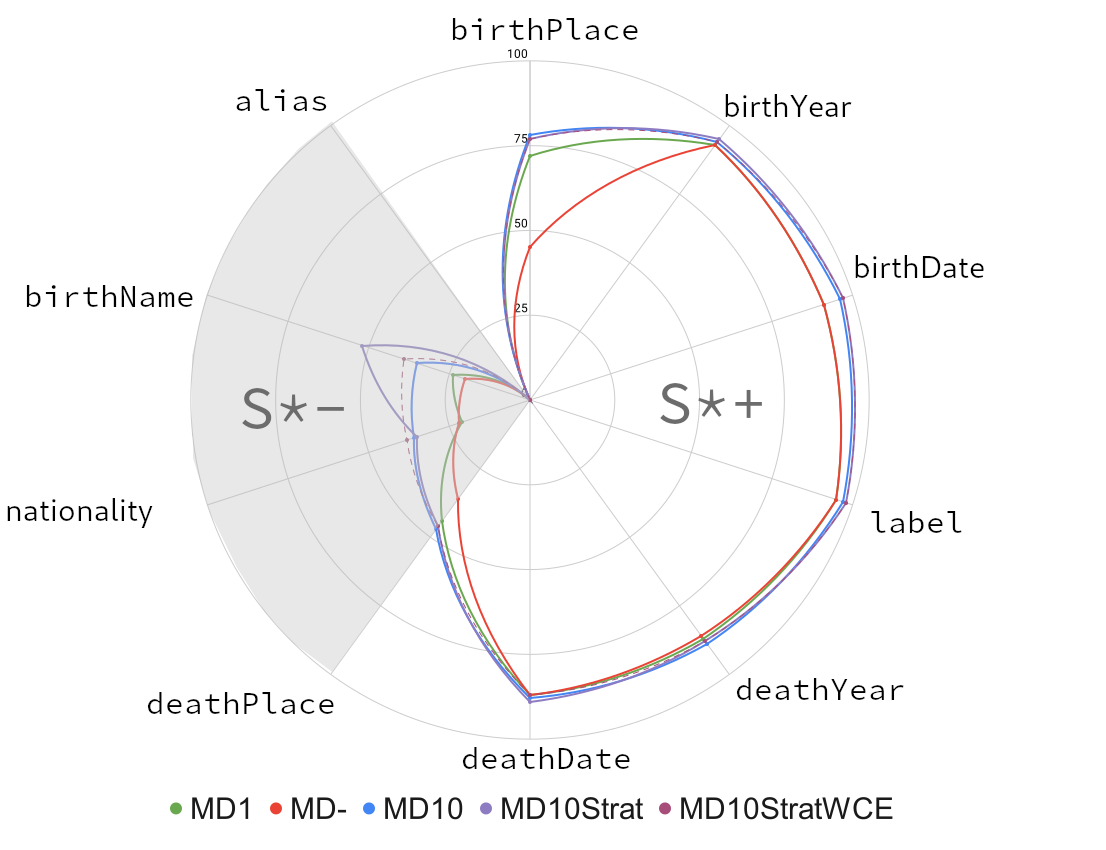}
        \caption{Evaluated on the random dataset $d_R$}
        \label{fig:DRstep1_2}
    \end{subfigure}
\end{figure}

Figure~\ref{fig:cross-dataset_comp} presents a property-level comparison of several models over the datasets $d_-$ and $d_R$.
First, we can see from the $d_-$ test set (Figure~\ref{fig:cross-dataset_comp}a) that scaling ($\mathcal{M}D10$) helps with rare properties, but is not sufficient to reach the same level as $\mathcal{M}D_-$ which was trained specifically with rare properties. Moreover, the stratified models ($\mathcal{M}D10_{Strat}$ and $\mathcal{M}D10_{StratWCE}$) under-perform on several properties compared to the simpler $\mathcal{M}D10$ model.
Concerning the random evaluation set ($d_R$, Figure~\ref{fig:cross-dataset_comp}b), we observe a slight improvement for \texttt{dbo:birthPlace}, \texttt{dbo:nationality} and \texttt{dbo:deathPlace}, but none of the models, including $\mathcal{M}D10$ and $\mathcal{M}D_-$, succeed in accurately extracting the rare \texttt{dbo:alias} property.
Concerning the frequent properties, we can see that all configurations perform well since $\mathcal{M}D1$ already performs well. In this context, scaling (e.g., $\mathcal{M}D10$) yields a modest performance gain.

\section{Reaching a Sufficient Exposure per Property}\label{sec:part3}

\subsection{Data Augmentation Strategies}
In the previous section, we showed that scaling and over-representation of rare properties can help improve model performance. However, the \texttt{dbo:alias} property remains particularly challenging. This property appears only 1,398 times in our distilled knowledge base (see Table 1) and has consistently been sampled below the sufficient exposure threshold of 1,000 examples, as emphasised by prior research~\cite{AL2024-knowledge3}. Such situations are common in low-resource scenarios. To address this, we propose to enrich our training data using three methods described in the next sections. Additional details of the data augmentation approaches in Sections~\ref{sec:DA2} and \ref{sec:DA3} are given in our Github repository\footnote{\url{https://github.com/datalogism/Kastor/blob/main/doc/AugStrategies.md}}.

\subsubsection{Using all the data available.}
We enrich dataset $D10$ by adding all the entities from the distilled knowledge base, that use the \texttt{dbo:alias} property. This creates a new dataset, $D10A_{all}$, which describes 13,244 distinct entities.

\subsubsection{Abstract-template knowledge replacement}\label{sec:DA2} 
We conduct two KG-pattern replacement strategies to augment $D10$.
Starting from the 156 entities whose graphs use the \texttt{dbo:alias} property, we derive 156 abstract templates that we use to build synthetic abstract-graph pairs and reach at least 1,000 examples using this property. Two strategies were conducted, both replacing graph data from a given entity with values from the abstract of another entity. 
They simply differ in the way entities are selected to populate the templates and they generate two datasets $D10A_{KR0}$, $D10A_{KR1}$ describing 12,928 and 13,200 entities, respectively, including multiple $(w,g)$ couples for the same entity in each dataset.

\subsubsection{Optimal sufficient exposure dataset}\label{sec:DA3} 
We construct a new balanced sample, $D4SE_{all}$, which comprises at least 1,000 examples for each property and encompasses 3,472 distinct entities. 
This process allows us to satisfy the sufficient exposure threshold defined for all the properties while minimizing the total size of the dataset. Concretely, $D4SE_{all}$ is built using random sampling without replacement that:

(1) sorts the properties of $s^*$ from the least to the most represented in $\mathcal{K}_{dist}(s^*)$;
(2) iterates over this list of properties, and selects new random abstract-graph pairs until reaching 1,000 couples per property. The process is stopped once all properties exceed the 1,000-example threshold.

\begin{table}[h!]
\caption{Number of triples per property of the new datasets}
\label{tab:dataset_stat_2}
\resizebox{\linewidth}{!}{
\begin{tabular}{l|l|r|r|rr|r}
           \multicolumn{2}{l|}{}   & $D10$ & $D10A_{all}$ & $D10A_{KR0}$ & $D10A_{KR1}$ & $D4SE_{all}$ \\ \hline 
\multirow{6}{*}{$s^*_+$} & birthYear   & 10,508 & 11,396        & 11,296        & 11,625        & 2,718         \\
                        & birthPlace  & 9,887  & 9,948         & 10,757        & 10,913        & 1,115         \\
                        & birthDate   & 9,553  & 10,387        & 10,305        & 10,450        & 2,361         \\
                        & label       & 8,211  & 9,158         & 8,989         & 9,277         & 2,321         \\
                        & deathYear   & 4,804  & 5,185         & 4,962         & 5,570         & 1,605         \\
                        & deathDate   & 4,237  & 4,544         & 4,387         & 4,877         & 1,353         \\ \hline 
\multirow{4}{*}{$s^*_-$} & deathPlace  & 3,181  & 3,197         & 3,279         & 3,777         & 1,024         \\
                        & nationality & 1,383  & 1,401         & 1,385         & 1,585         & 1,001         \\
                        & birthName   & 1,222  & 1,434         & 1,296         & 1,728         & 1,001         \\
                        & alias       & \underline{156}   & 1,398         & 1,082         & 1,387         & 1,001         \\

\end{tabular}
}
\vspace{-5mm}
\end{table}
\subsection{Results}
The training details for the models referred to in this section are available on Wandb\footnote{Model training: \url{https://wandb.ai/celian-ringwald/GenLimitsPart3_train},\\ Cross-Evaluation:\url{https://wandb.ai/celian-ringwald/GenLimitsPart3_crossEval}}.
To evaluate the produced models, we reused the cross-evaluation set introduced in Section~\ref{sec:part2}. Building on insights from~\cite{DBLP:conf/esws/RingwaldGFMA25}, we also extended our analysis with a corrected evaluation set: $d_C$. This dataset helps assess whether errors observed in $d_R$ stem from the discovery of new relevant facts (in the case of false positives), from omissions, or from noise in the KG (in the case of false negatives). We manually annotated 1,470 errors—comprising all false positives (FP) and false negatives (FN) produced during the evaluation of $d_R$ by the models $\mathcal{M}D1$, $\mathcal{M}D\_$, $\mathcal{M}D10$, but also $\mathcal{M}D10A_{all}$, $\mathcal{M}D10A_{KR0}$, $\mathcal{M}D10A_{KR1}$ and $\mathcal{M}D4SE_{all}$. 

In Table~\ref{tab:Res_all_2}, considering the $d_C$ results which are the ones that matter most, we note that $\mathcal{M}D4SE_{all}$ outperforms all the other configurations by 10 points of F1. All the models were previously overestimated when evaluated solely on $d_R$. 
The results recorded by $D10A_{all}$, $\mathcal{M}D10A_{KR0}$, $\mathcal{M}D10A_{KR1}$ are close to the original $\mathcal{M}D10$ model.

\begin{figure}[p]
    \centering
    \caption{Averaged micro F1 by property and model}
    \label{fig:DC_compare}

    \begin{subfigure}[b]{0.8\linewidth}
        \centering
        \includegraphics[width=\linewidth, height=0.27\textheight, keepaspectratio]{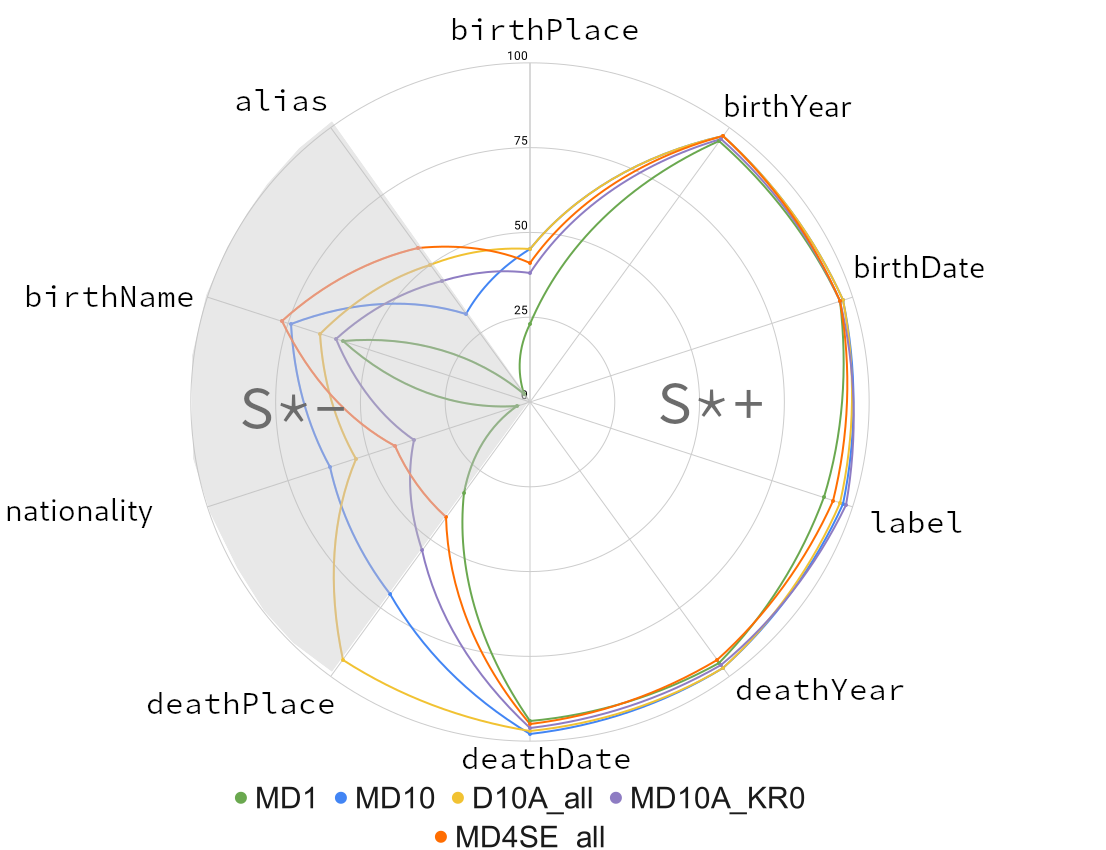}
        \caption{Evaluated on the rare properties dataset $d_-$}
        \label{fig:D_-_step2_2}
    \end{subfigure}

    \vspace{0.5em}

    \begin{subfigure}[b]{0.8\linewidth}
        \centering
        \includegraphics[width=\linewidth, height=0.27\textheight, keepaspectratio]{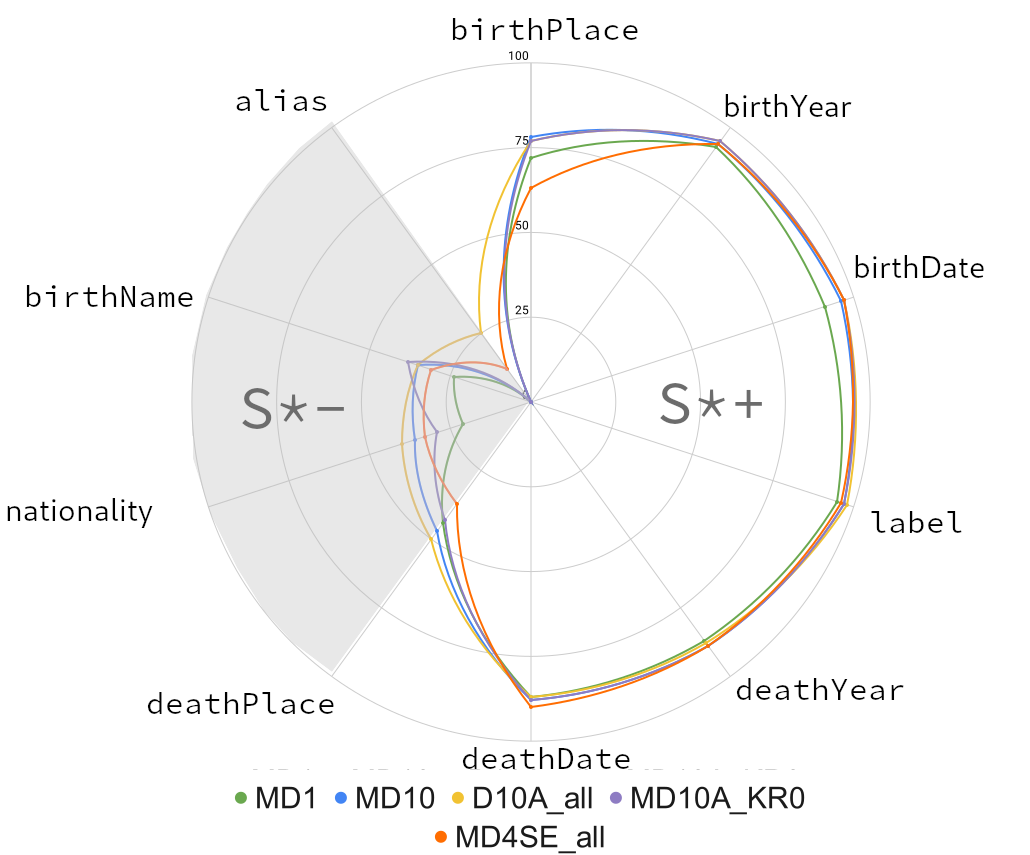}
        \caption{Evaluated on the random dataset $d_R$}
        \label{fig:R_step2_2}
    \end{subfigure}

    \vspace{0.5em}

    \begin{subfigure}[b]{0.8\linewidth}
        \centering
        \includegraphics[width=\linewidth, height=0.27\textheight, keepaspectratio]{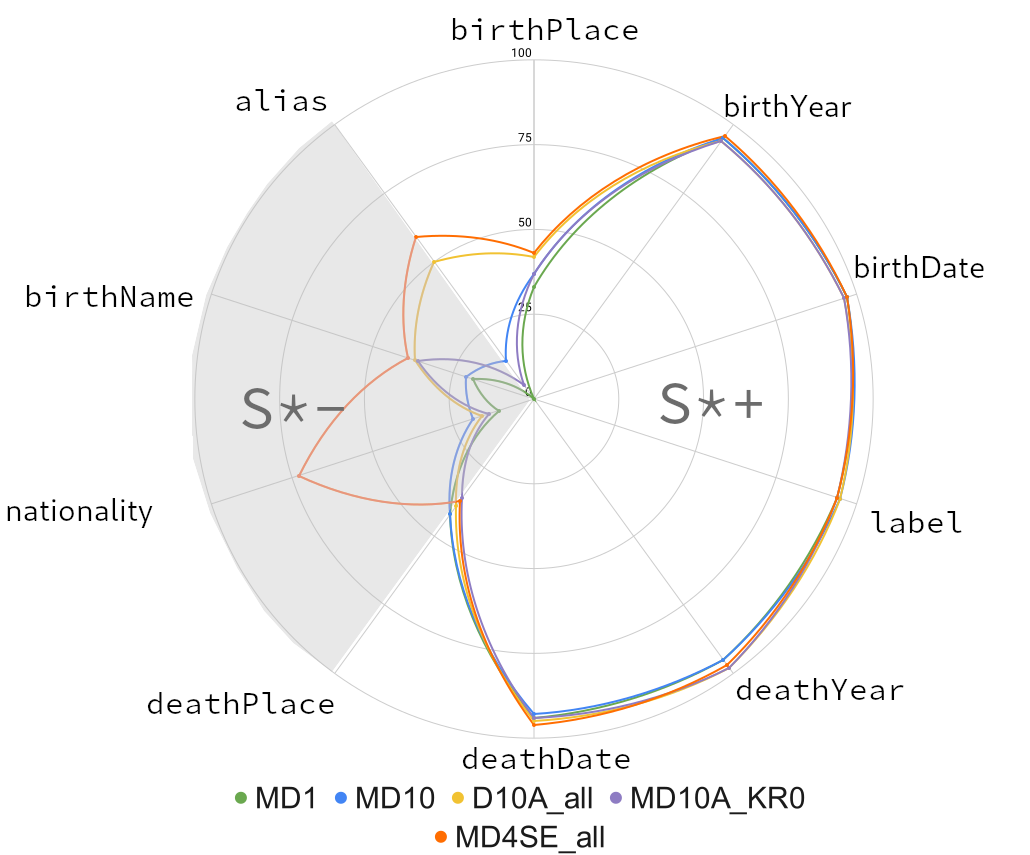}
        \caption{Evaluated on the corrected dataset $d_C$}
        \label{fig:DC_step2_2}
    \end{subfigure}
\end{figure}

We also analyse per-property model performance  on a selection of evaluation datasets, as shown in Figure~\ref{fig:DC_compare}.  
Firstly, Figure~\ref{fig:D_-_step2_2} shows that on dataset $d_-$, $\mathcal{M}D4SE_{all}$ and $\mathcal{M}D10A_{all}$ slightly outperform the other models on some rare properties: \texttt{dbo:deathPlace}, \texttt{dbo:birthName}, \texttt{dbo:alias}. 
Secondly, Figure~\ref{fig:R_step2_2} shows that on dataset $d_R$, $\mathcal{M}D4SE_{all}$ and $\mathcal{M}D10A_{all}$ are the only configurations capable of extracting \texttt{dbo:alias}, which was specifically the property underrepresented in previous scenarios. 
Finally, the results recorded on $d_C$ differ notably from $d_R$. Indeed, Figure~\ref{fig:DC_step2_2} shows that many of the errors produced by $\mathcal{M}D10A_{all}$ and $\mathcal{M}D4SE_{all}$ were in fact genuine knowledge discoveries. This makes $\mathcal{M}D4SE_{all}$ the most robust model overall, particularly from the perspective of handling rare properties, such as \texttt{dbo:alias}.

\begin{table*}\caption{Experimental results on each model variation with the five proposed cross-evaluation sets. Metrics are averaged over all the folds. The standard deviation is always less than 5pt. Best scores are in bold and worst ones are underlined.}\label{tab:Res_all_2}
\resizebox{\linewidth}{!}{
\begin{tabular}{l|rrlrr|lllll}
\multirow{2}{*}{Model}  & \multicolumn{5}{c|}{$F_1^-$}                                                                                            & \multicolumn{5}{c}{$F_1^+$}                                \\ \cline{2-11} 
                        & \multicolumn{1}{l}{$d_N$} & \multicolumn{1}{l}{$d_+$} & $d_-$ & \multicolumn{1}{l|}{$d_R$} & \multicolumn{1}{l|}{$d_C$} & $d_N$ & $d_+$ & $d_-$ & \multicolumn{1}{l|}{$d_R$} & $d_C$ \\ \hline
$\mathcal{M}D10(PLAIN, s^*)$        & 84.42                     & \textbf{91.94}                     & \textbf{86.05} & \multicolumn{1}{r|}{86.81} & 72.46                      & 67.05 & 97.05 & 77.26 & \multicolumn{1}{l|}{66.52} & 60.97 \\
$\mathcal{M}D10A_{all}(PLAIN, s^*)$ & 84.00                     & 91.85                     & 85.15 & \multicolumn{1}{r|}{\textbf{87.19}} & 72.36                      & 67.44 & 96.38 & 76.65 & \multicolumn{1}{l|}{69.54} & 63.92 \\
$\mathcal{M}D10A_{KR0}(PLAIN, s^*)$ & 84.00                     & 91.70                     & \underline{82.07} & \multicolumn{1}{r|}{86.71} & \underline{71.29}                      & \underline{64.57} & 95.72 & \underline{71.22} & \multicolumn{1}{l|}{65.49} & \underline{58.26} \\
$\mathcal{M}D10A_{KR1}(PLAIN, s^*)$ & \textbf{86.85}                     & 91.85                     & 85.02 & \multicolumn{1}{r|}{86.85} & 72.47                      & \textbf{67.30} & \textbf{97.11} & 71.22 & \multicolumn{1}{l|}{\textbf{85.02}} & 59.64 \\
$\mathcal{M}D4SE_{all}(PLAIN, s^*)$ & \underline{76.41}                     & \underline{79.05}                     &  85.69 & \multicolumn{1}{r|}{\underline{79.66}} & \textbf{79.90}                      & 65.07 & \underline{89.23} & \textbf{89.24} & \multicolumn{1}{l|}{\underline{64.62}} & \textbf{72.74}
\end{tabular}
}
\end{table*}

\section{Discussion}
Our experiments provide evidence of the ability of our models to extract RDF patterns from a single SHACL shape containing both datatype and object properties, with high performance when it comes to extract frequent properties. Moreover, we have also shown that our models are not affected by memorisation issues. Our work also highlighted that smaller and more specialised SLMs could outperform general models like REBEL, by requiring less training resources.  
Concerning the input text type, we showed that the markdown text did not bring any improvement over the plain text. 
A future work could explore metrics specifically tailored to object property comparison to help us evaluate how close the generated triples are from the expected answer, instead of using a strict equality criteria.
We observed that the stratification strategy and the proposed weighted loss did not significantly improve performance in contexts with unbalanced property distributions. This limitation can be attributed to the simplicity of our designs: 
(1) stratification focused only on a limited subset of properties, while in reality graphs are associated with various properties;
(2) the weight was applied solely at the sequence level. 
Finally, we highlighted the efficiency of methods when we ensure sufficient exposure for each property during training. We showed that using a reduced dataset containing at least 1,000 examples per property enables the model to perform consistently well across various properties and test scenarios. 

But while we emphasise the availability of a minimum number of labelled examples, the naive KG-based augmentation methods proposed in this work proved to be inefficient, primarily due to their inherent limitations—namely, the introduction of textual inconsistencies and the generation of multiple, redundant inputs for the same entity. 
Nonetheless, future extensions of our approach should focus on the design of more advanced and semantically consistent data augmentation techniques, specifically tailored for low-resource environments

\section{Conclusion}
In this paper we extended our RDF pattern-based RE approach to the extraction of both datatype and object properties. We also highlighted the fact that the overall performance of SLMs finetuned for RE is highly sensitive to the property distribution in the training set: they perform well on frequent properties but struggle significantly with rare properties.
To tackle this issue, we proposed and tested different strategies, among which only scaling helps increase performance at the micro level, without filling the gap recorded at the macro level. To fill this gap, we showed that the model should be trained by respecting a minimal amount of exposure to properties, in order to perform better and equally among the frequent and rare properties. As future work, the highlighted challenges related to data augmentation should be considered to generalise our results to a broader set of SHACL shapes.

\paragraph{\textbf{Acknowledgments}}
This work has been supported by the 3IA Côte d'Azur Investments (ANR-23-IACL-0001), the UCAJEDI (ANR-15-IDEX-01), the OPAL infrastructure, and Université Côte d’Azur’s Center for High-Performance Computing. 
Grammar was improved with ChatGPT (GPT-4) and Grammarly, and all AI-generated content was thoroughly reviewed, edited, and validated by the authors, who take full responsibility for the final manuscript and all its content.

\bibliographystyle{ACM-Reference-Format}
\bibliography{biblio}

\end{document}